\documentclass[11pt]{article}

\usepackage[final]{acl}

\usepackage{times}
\usepackage{latexsym}

\usepackage[T1]{fontenc}

\usepackage[utf8]{inputenc}

\usepackage{microtype}

\usepackage{inconsolata}

\usepackage{graphicx}
\usepackage{booktabs}

\title{Agent Architecture Cost–Success Trade-offs: A Controlled Evaluation across Enterprise-Style SOPs}

\author{
  Subhrangshu Nandi\textsuperscript{1*}, Hitesh Arora\textsuperscript{1*}, Yunfei Bai\textsuperscript{1*}, Udita Patel\textsuperscript{1*}, Chirag Shah\textsuperscript{1,2*}
  \AND
  Hannah Kwon\textsuperscript{1}, Wei Zhang\textsuperscript{1}, Weian Chen\textsuperscript{1}, Chloe Ruoqi Li\textsuperscript{1}, Alberto Sconfienza\textsuperscript{1}
  \AND
  Ayush Goyal\textsuperscript{1}, Yasmine Lee\textsuperscript{1}, Ming Xiong\textsuperscript{1}, Prathap Ramachandra\textsuperscript{1}
  \AND
  {\normalfont \textsuperscript{1}Amazon \quad \textsuperscript{2}University of Washington}
}

\begin{document}
\maketitle
{\renewcommand{\thefootnote}{*}\footnotetext{Equal contribution.}}
{\renewcommand{\thefootnote}{}\footnotetext{Correspondence: \href{mailto:hitesaro@amazon.com}{\texttt{hitesaro@amazon.com}}}}

\begin{abstract}
Enterprises increasingly use large language model agents to execute standard
operating procedures (SOPs): long, branching, tool-intensive workflows. Yet there
is little controlled evidence for choosing among competing agentic architectures.
Holding the model, tasks, and tools fixed and varying only the architecture, we
compare four designs (a function-calling single agent, a skill-augmented agent,
centralized multi-agent orchestration, and a triage swarm; eight arms) across 13 enterprise
SOPs, and introduce a skill-definition quality rubric to build matched high- and
low-quality skill corpora. Three takeaways emerge. (1)~\textbf{There is no universal
best architecture; the right choice is SOP-dependent.} On a given SOP, task success
ranges from under $10\%$ on the hardest to near $100\%$ on the easiest; no
architecture is statistically separable in aggregate, and added structure only
plausibly pays off on complex, branching SOPs (a directional, not yet significant,
trend). (2)~\textbf{Added architecture cost must be justified by
a task-success gain.} Every coordinated arm (multi-agent or skill-augmented) spends
$1.4$--$4.7\times$ the tokens and $2.6$--$5.8\times$ the latency of the single agent
for gains that are usually small and rarely significant, making the single agent the
efficient default; high-quality
skills do lift success (to $73.1\%$ from $67.4\%$) while degraded skills fall below
it. (3)~\textbf{Implementation quality outweighs the single- versus multi-agent
choice.} A plausible multi-agent implementation loses to the single agent, yet a
targeted redesign of it becomes the only arm to significantly beat the baseline
($75.6\%$ vs.\ $67.4\%$).
\end{abstract}

\begin{figure}[!t]
\centering
\includegraphics[width=0.85\columnwidth]{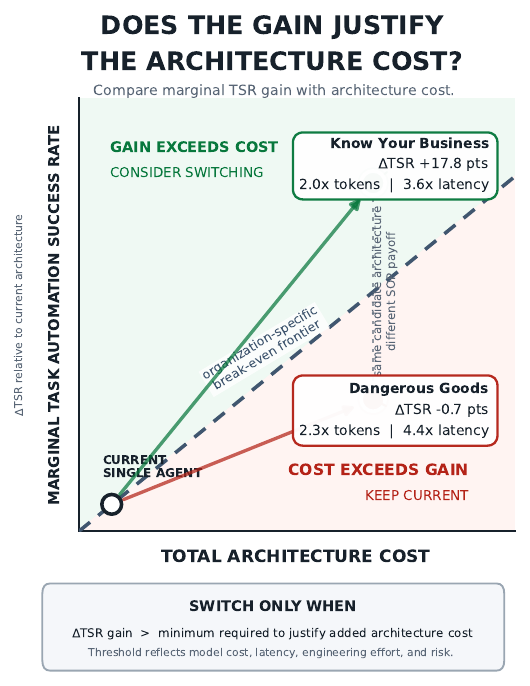}
\caption{Architecture selection as an SOP-specific cost--benefit decision.
A candidate should replace the current architecture only when its marginal gain
in task success rate is sufficient to justify its total architecture cost. The
callouts show measured changes from the function-calling baseline to the
descriptive swarm on two illustrative SOPs; the dashed break-even frontier is
conceptual and organization-specific, not estimated from these data. This figure
is for illustrative purposes only.}
\label{fig:teaser}
\end{figure}
\vspace{-0.5cm}

\vspace{-0.2cm}
\section{Introduction}
Enterprises increasingly rely on large language model (LLM) agents to automate
long, multi-step standard operating procedures (SOPs): branching, tool-intensive
workflows in which an agent must interpret instructions, select and parameterize
tools, carry state across dependent steps, and commit to a decision. Several agentic
architectures address this: a \emph{function-calling} single agent, a
\emph{skill-augmented} agent that loads reusable skills on demand
\citep{li2026skillsbench,li2026clawsbench}, \emph{centralized orchestration} that
decomposes the procedure for workers with isolated context
\citep{wang2023planandsolve,xu2023rewoo,kim2024llmcompiler}, and a
\emph{triage-guided swarm} of peer specialists that hand off control
\citep{liu2024agentpattern}. Yet practitioners have little controlled evidence for
choosing among them: prior work studies skills or coordination mechanisms in
isolation \citep{li2026skillsbench,chen2026scalinglaws,li2026single}, and none holds
the model, tasks, and tools fixed while varying only the architecture on long,
tool-intensive SOPs.

We close this gap with SOP-Bench \citep{nandi2026sopbench}, whose deterministic
mock-tool dispatch fixes tool outcomes so that observed differences reflect the
architecture and skill design rather than external services. Holding the model
(\texttt{claude-opus-4-8}), tasks, and tools fixed, we compare the four architectures
as eight arms across 13 SOPs; the skill and swarm arms each
carry an internal knob, and multi-agent orchestration is evaluated in two implementations. We
additionally contribute the \emph{Skill-Definition Quality Rubric}, a
multi-dimensional instrument that scores a skill as a reusable procedural artifact
rather than a bundle of tool schemas, and use it to build matched high- and
low-quality skill corpora. Figure~\ref{fig:teaser} frames the practitioner decision
the study informs: weigh an architecture's marginal accuracy gain against its added
cost.

Three findings follow. First, the SOP complexity dominates the architecture: per-SOP success
ranges from single digits to near-ceiling and no architecture is statistically
separable in aggregate, so we treat the SOP as the unit. Second, cost, not accuracy, separates the architectures, making the
single agent the efficient default. Third, implementation quality outweighs the single- versus multi-agent choice: a
poorly built orchestrator loses to the single agent, while a targeted fix makes it
the only arm to significantly beat the baseline.

\section{Related Work}
\label{sec:related_work}
We review two strands of literature here that closely relate to our work.

\paragraph{Single- vs.\ multi-agent designs.}
Single-agent methods couple reasoning to tool use (ReAct~\cite{yao2023react},
Toolformer~\cite{schick2023toolformer}) and increasingly package reusable
capabilities as \emph{skills} loaded on demand, consolidated in industry under the
Model Context Protocol~\cite{anthropic2024mcp} and Agent Skills~\cite{anthropic2025skills}.
Multi-agent systems instead distribute work~\cite{liu2024agentpattern}, through
either centralized orchestration (AutoGen~\cite{wu2023autogen},
MetaGPT~\cite{hong2023metagpt}) or decentralized, handoff-based swarms (OpenAI
Swarm~\cite{openai2024swarm}, GPTSwarm~\cite{zhuge2024gptswarm}). Added agents are
not free: they can beat a single agent but at roughly an order of magnitude more
tokens~\cite{anthropic2025multiagent}, frequently underperform for identifiable
reasons~\cite{cemri2025mast}, and a single skill-equipped agent can sometimes match
them~\cite{li2026single}. This motivates our question of when orchestration or swarm
designs justify their overhead.

\paragraph{Evaluating agents and skills.}
Agent evaluation has moved toward execution-grounded outcomes~\cite{liu2023agentbench,zhou2023webarena,jimenez2023swebench},
tool-call correctness~\cite{patil2024bfcl}, reliability across repeated
trials~\cite{yao2024taubench}, and process-level judging~\cite{ma2024agentboard}. A
fast-emerging line analyzes skills as first-class objects: effectiveness and
scaling~\cite{li2026skillsbench,li2026clawsbench,chen2026scalinglaws}, quality and
representation~\cite{gao2026skillreducer,liang2026ssl}, selection at
scale~\cite{li2026agentskillos}, continual learning~\cite{zhong2026skilllearnbench},
and security~\cite{wu2026skillscope,zhuang2026agenttrap}. Prior work isolates these
dimensions but offers no controlled, cost-normalized comparison of skill-augmented
single agents against orchestration and swarm designs on shared, procedure-level
tasks (SOP-Bench~\cite{nandi2026sopbench}). We fill this gap.

\section{Methodology}
\label{sec:methods}
This section provides details about how we designed experiments and evaluated various arms.

\subsection{Experimental Setting}
\label{sec:setting}

The primary objective of our work is to investigate how the agentic architecture and procedural-skill design affect the execution
of enterprise standard operating procedures (SOPs). For evaluation, we use SOP-Bench
\citep{nandi2026sopbench}, whose tasks pair natural-language
procedures and tool schemas with deterministic mock-tool dispatch and
ground-truth outputs. Deterministic dispatch fixes tool outcomes across arms, so
observed differences reflect the agentic architecture and skill design rather than
variation in external services. 

\paragraph{Architectures, arms, and model.}
We compare four agentic architectures spanning eight arms, summarized in
Table~\ref{tab:arms}: a function-calling baseline, a skill-augmented single agent
(Constrained Skill Execution, CSE), centralized orchestration, and a triage-guided
swarm. For the skill and swarm
families we designate one representative arm ($\dagger$ in Table~\ref{tab:arms};
CSE-high, swarm-descriptive) for cross-family statements and report the
alternatives separately; the two orchestrator implementations are reported as
distinct arms. All cross-architecture
runs hold the model fixed at \texttt{us.anthropic.claude-opus-4-8} on a pinned code
revision. We fix the version after confirming the baseline's macro TSR varies by at
most $2.3$ points across Opus 4.6, 4.7, and 4.8 (well within the between-SOP
standard errors; Appendix~\ref{sec:appendix-baseline-models}), so the choice
isolates architecture effects without favoring any arm. Because all runs use one
model family, we make no claim that our findings generalize across families. The shared comparison spans the 13
SOPs common to all arms (2{,}124 tasks per arm). 

\begin{table*}[t]
\centering
\small
\begin{tabular}{@{}llp{9.2cm}@{}}
\toprule
\textbf{Architecture family} & \textbf{Arm} & \textbf{Distinguishing setting} \\
\midrule
Function calling & Baseline & Single agent; receives the full SOP and all declared tools. \\
\addlinespace
Skill-augmented (CSE) & CSE-high$^\dagger$ & High-quality skills; executor exposed only to each skill's declared tools. \\
                      & CSE-low  & Degraded skills; same tool scoping. \\
\addlinespace
\parbox[t]{3cm}{Multi-agent (centralized orchestration)} & Orch-v1  & Decomposer--worker--synthesizer; each worker sees an isolated SOP slice and tool subset. \\
                          & Orch-v2  & Full SOP and tool set per worker; commit-with-basis contract, one re-dispatch round, output validation. \\
\addlinespace
Triage-guided swarm & Swarm-none & No shared procedure blueprint. \\
                    & Swarm-descriptive$^\dagger$ & Common goal and unordered high-level stages (no tool-level instructions or ordering). \\
                    & Swarm-full & Recommended stage order and routing logic (hybrid: central planning, decentralized execution). \\
\bottomrule
\end{tabular}
\caption{Agentic architectures and the eight arms of the shared comparison, all run
on a single fixed model (\texttt{claude-opus-4-8}) and pinned code revision.
$\dagger$ marks the representative arm used for cross-family statements. Open-Tool
Skill Guidance (OSG), a separate skill formulation evaluated on only two SOPs, is
described in \S\ref{sec:agent-configurations} and is not part of the shared matrix.}
\label{tab:arms}
\end{table*}

\subsection{Agent Configurations}
\label{sec:agent-configurations}

Table~\ref{tab:arms} summarizes the eight arms; Appendix~\ref{sec:appendix-configs}
gives full prompts and control flow. The \emph{function-calling baseline} is a
single agent with the full SOP and all tools. \emph{Constrained Skill Execution}
(CSE) selects skills per task and exposes the executor only to each selected
skill's declared tools, testing skills as both guidance and hard capability
boundaries. \emph{Open-Tool Skill Guidance} (OSG), a separate formulation evaluated
on two complex SOPs, supplies the same skills as advice but exposes all tools,
logging out-of-scope calls as deviations. \emph{Centralized orchestration} runs a
decomposer, workers, and a synthesizer, each worker seeing only its SOP slice and
tool subset (orchestration-v1); \emph{orchestration-v2} instead gives every worker
the full SOP and tool set, forbids placeholder outputs, and adds one re-dispatch
round and output validation. The \emph{triage-guided swarm} routes among
tool-scoped specialists over a shared blackboard, ablated at three blueprint levels
(none, descriptive, full).

\subsection{Skill-Definition Quality Rubric}
\label{sec:skill-quality-rubric}

A contribution of this work is a rubric that scores a skill \emph{definition} as a
reusable procedural artifact rather than a bundle of tool schemas: scoring tool
declarations alone would miss the procedural guidance, routing cues, constraints,
and disambiguation signals that determine how a planner discovers, selects, and
executes a skill. Multi-dimensional agentic-evaluation frameworks
\citep{mehta2025beyond} and recent skill-evaluation efforts
\citep{ye2026claw,chen2026scalinglaws} likewise argue that richer, structured
assessment tracks downstream behavior more faithfully than a single accuracy score.
Building on this, the rubric assigns 50 points across five dimensions (identity and
discoverability, instruction quality, contract clarity, disambiguation signals, and
governance metadata), scored by an LLM judge under a two-stage structural-then-semantic
protocol. Appendix~\ref{sec:appendix-rubric} gives the point allocations, grade bands,
protocol, and worked high- and low-quality examples.

The rubric scores definition quality only, not execution correctness or task
success. We use it to build and verify the two skill corpora rather than as an
outcome measure: the high-quality corpus scores near the ceiling (pooled mean
$99.8\%$) and the degraded corpus far lower ($67.6\%$), a large and consistent
separation. This is a manipulation check on our intervention, not evidence that
rubric score predicts task success, a claim we do not make. The low quality corpus is a
systematic degradation of the high quality one (preserving identity, tool dependencies, and
workflow structure) used for both CSE and OSG, so the skill-quality contrast is a
bundled manipulation of artifact richness, not the causal effect of a single field.

\vspace{-0.3cm}
\subsection{Evaluation and Analysis}
\label{sec:evaluation}

We parse each agent's final output and compare it to ground truth with a canonical
fuzzy-matching parser (all fields must match for multi-field tasks), and report two
metric families: \textbf{outcome} (task success rate,
$\mathrm{TSR}=\#\mathrm{correct}/\#\mathrm{tasks}$, over the full task count) and
\textbf{cost} (tokens, wall-clock latency, and LLM calls as the
architecture-neutral effort unit).

We treat the SOP as the unit of inference: with 13 SOPs the between-SOP (macro)
standard errors (roughly 7 to 9 points) dominate the pooled task-level errors (about
1 point), so we report per-SOP TSRs with binomial standard errors as primary
evidence, pair every pooled $p$-value with a per-SOP sign test, and do not read
pooled significance as a general ranking. For a confirmatory single-arm-versus-baseline
test we fit a random-effects model over the 13 per-SOP log-odds ratios
(DerSimonian--Laird with the Knapp--Hartung small-sample adjustment;
Appendix~\ref{sec:appendix-randomeffects}).

\section{Results}
\label{sec:results}

\subsection{The SOP dominates the architecture}
\label{sec:results-sop}

Table~\ref{tab:main} reports per-SOP TSR for all eight arms, with macro TSR and its
between-SOP standard error in the final row; we treat the per-SOP rate as the unit
of inference (\S\ref{sec:evaluation}).

Within every arm the per-SOP TSR spans nearly the full range, from single digits on
the hardest SOPs (\texttt{content\_flagging}, \texttt{video\_annotation}) to
near-ceiling on the easiest (\texttt{referral\_abuse}, \texttt{email\_intent}); SOP
identity explains far more variance than the agentic architecture. The between-SOP
(macro) standard errors are large, $6.3$--$7.8$ points, and overlap across all arms;
because this uncertainty is between-SOP, more tasks per SOP do not tighten it.
Consequently no arm separates from the baseline by a per-SOP sign test, and the
macro ordering in Table~\ref{tab:main} cannot be asserted as a general ranking, with
one defensible exception isolated by a paired test in \S\ref{sec:results-orch}.

\begin{table*}[t]
\centering
\setlength{\tabcolsep}{4pt}
\resizebox{\textwidth}{!}{%
\begin{tabular}{@{}lrrrrrrrr@{}}
\toprule
SOP & FC & CSE-hi & CSE-lo & Orch-1 & Orch-2 & Sw-none & Sw-desc & Sw-full \\
\midrule
Aircraft & 22 & 44 (3.5,2.6) & 49 (2.4,2.4) & \textbf{52} (1.4,2.0) & 46 (4.3,2.7) & 46 (2.2,3.4) & 48 (2.6,4.1) & 48 (2.4,9.3) \\
Content flag. & 29 & 37 (1.9,1.7) & \textbf{38} (1.6,2.0) & 29 (1.0,1.4) & 33 (3.3,2.0) & 31 (2.0,3.5) & 35 (1.6,3.3) & \textbf{38} (2.0,3.2) \\
Customer service & 88 & 81 (2.4,2.9) & 74 (2.1,3.2) & 74 (1.3,4.0) & 90 (5.5,5.2) & \textbf{94} (1.3,3.9) & \textbf{94} (1.4,4.1) & \textbf{94} (1.5,4.2) \\
Dangerous goods & \textbf{92} & 77 (1.8,1.4) & 90 (2.4,2.2) & 75 (1.8,2.5) & 87 (4.4,2.0) & 88 (2.3,5.0) & 91 (2.3,4.4) & 90 (2.7,4.2) \\
Email intent & 87 & 91 (13.8,5.5) & 52 (8.7,4.4) & 76 (2.7,3.0) & \textbf{96} (8.9,4.1) & 91 (2.3,4.3) & 90 (2.4,4.7) & 91 (2.7,4.8) \\
KYB & 39 & \textbf{86} (3.2,2.9) & 40 (2.8,2.6) & 23 (1.8,2.3) & 60 (4.0,2.2) & 53 (2.4,3.4) & 57 (2.0,3.6) & 57 (3.5,10.6) \\
Patient intake & 94 & 94 (2.5,1.7) & 94 (1.7,1.6) & 14 (1.0,1.6) & \textbf{98} (3.1,1.3) & 56 (2.5,3.9) & 74 (2.4,3.6) & 79 (2.9,3.4) \\
Referral v1 & 84 & \textbf{97} (2.2,2.1) & 96 (1.7,1.7) & 96 (2.1,3.0) & 95 (4.3,3.3) & 93 (2.3,3.8) & 93 (2.1,3.8) & 93 (2.6,3.7) \\
Referral v2 & \textbf{100} & 94 (4.5,3.2) & 2 (3.8,3.3) & 99 (2.2,3.1) & \textbf{100} (6.8,3.6) & 72 (4.0,7.6) & 98 (4.0,7.1) & 99 (8.9,9.6) \\
Traffic spoof. & 76 & 78 (4.0,3.5) & 78 (2.8,3.2) & \textbf{80} (2.4,3.2) & \textbf{80} (4.5,3.6) & 78 (3.5,6.7) & 77 (3.2,5.4) & 76 (3.9,5.2) \\
Video annot. & 34 & 38 (1.3,1.8) & 34 (1.0,1.5) & 31 (1.0,3.3) & \textbf{62} (6.6,4.6) & 34 (1.6,4.0) & 37 (0.9,3.3) & 35 (2.0,5.4) \\
Video class. & 79 & \textbf{85} (3.8,3.8) & 80 (3.5,3.8) & 58 (1.0,2.6) & \textbf{85} (3.3,3.5) & 84 (1.9,4.8) & 82 (1.7,4.3) & 82 (2.1,4.1) \\
Warehouse & \textbf{52} & 50 (4.9,4.5) & 51 (3.1,3.6) & 51 (1.8,2.4) & 50 (4.6,2.9) & 30 (1.9,3.7) & 41 (1.9,4.1) & 35 (2.3,4.0) \\
\midrule
Macro $\pm$ SE & 67.4$\pm$7.7 & 73.1$\pm$6.3 & 59.8$\pm$7.8 & 58.2$\pm$7.7 & 75.6$\pm$6.3 & 65.4$\pm$6.9 & 70.4$\pm$6.6 & 70.5$\pm$6.8 \\
\bottomrule
\end{tabular}}
\caption{Per-SOP task success rate (TSR, \%) for all eight arms, with each coordinated cell annotated \emph{TSR (tokens$\times$, latency$\times$)}: the token and wall-clock cost of that arm on that SOP as a multiple of the function-calling baseline (FC) on the \emph{same} SOP. FC cells show TSR only (the $1.0\times$ reference); \textbf{bold} marks the highest-TSR arm on each SOP (all tied arms bolded). The final row is the macro TSR (mean over the 13 SOPs) $\pm$ between-SOP SE. Because multiples are computed per SOP, a column mean differs from the pooled figures in Table~\ref{tab:cost}.}
\label{tab:main}
\end{table*}

\subsection{Single-agent baseline versus orchestration}
\label{sec:results-orch}

The function-calling baseline beats orchestration-v1 on $9$ of $13$ SOPs
($6$ significant at $p<0.05$, two-proportion), the strongest single- versus
multi-agent result in the study, and v1 posts the lowest macro TSR of any arm
($58.2\%$). Its failures concentrate on coupled, multi-field SOPs
(\texttt{patient\_intake} $13.6\%$, \texttt{know\_your\_business} $23.3\%$) and
trace to a decomposition gap, not decomposition itself: each worker sees only its
slice and none may resolve a missing input, so the pipeline abstains with
placeholder decisions (\S\ref{sec:agent-configurations}). Closing that gap changes
the verdict: orchestration-v2 recovers exactly those SOPs
(\texttt{patient\_intake} $13.6\rightarrow98.5$, \texttt{email\_intent}
$75.8\rightarrow96.2$, \texttt{know\_your\_business} $23.3\rightarrow60.0$) and is
the only arm to significantly exceed the baseline (paired random-effects on per-SOP
log-odds, odds ratio $1.72$, $p=0.007$). We read this not as ``multi-agent beats
single-agent'' but as implementation quality mattering more than the single- versus
multi-agent choice: the same family loses when workers abstain and wins once they
must commit.

\subsection{Skill-quality and swarm-blueprint knobs}
\label{sec:results-knobs}

\paragraph{Skill quality helps, but only on a few SOPs.}
Higher skill quality raises macro TSR by $13.3$ points (CSE-high $73.1\%$ vs.\
CSE-low $59.8\%$), but the gap concentrates in a few long, branching procedures,
most sharply \texttt{know\_your\_business} ($+45.6$) and \texttt{email\_intent}
($+39.2$); elsewhere the corpora are within noise, and the difference is not
significant by a per-SOP sign test ($7$ wins vs.\ $4$, $2$ ties). The degraded
corpus even falls below the single-agent baseline ($59.8\%$ vs.\ $67.4\%$): where
skills are poor, the added selection-and-execution machinery hurts more than it
helps, sometimes silently (on referral\_v2 the degraded skill drops the SOP's output
contract, so near-correct decisions are filed under a concocted key and scored
wrong; Appendix~\ref{sec:appendix-cselow-v2}). Per-SOP results are in
Appendix~\ref{sec:appendix-skillquality}.

\paragraph{A swarm blueprint helps; more structure is just cost.}
A shared blueprint lifts macro TSR from $65.4\%$ (\texttt{none}) to $70.4\%$
(\texttt{descriptive}), but a rigid, ordered prescription adds nothing:
\texttt{full} ($70.5\%$) and \texttt{descriptive} are statistically
indistinguishable (sign test $p=1.00$) while \texttt{full} costs $42\%$ more tokens
and $39\%$ more latency (\S\ref{sec:results-cost}). The gain over \texttt{none} is
stage coverage, not smarter agents: without a blueprint, peers hand off before all
specialists run, so on referral-abuse v2 the bare swarm over-escalates $55$ of $200$
tasks (missing indicators) that both blueprinted arms get right. \texttt{full}'s
``all stages must run'' rule removes this but over-broadly, overriding legitimate
early-exit gates (on know-your-business it pushes past an incomplete-documentation
stop and flips to escalate), which is why it buys no accuracy over the descriptive
map.

\subsection{Cost is the robust separator}
\label{sec:results-cost}

Accuracy differences are mostly small and insignificant
(\S\ref{sec:results-sop}); cost differences are large, robust, and independent of
answer correctness. Table~\ref{tab:cost} shows every coordinated arm spends
$1.4$--$4.7\times$ the baseline's tokens and $2.6$--$5.8\times$ its latency.

\begin{table}[t]
\centering
\small
\begin{tabular}{lrrr}
\toprule
Arm & TSR & Tokens/task & Latency \\
    & (\%) & ($\times$base) & ($\times$base) \\
\midrule
Baseline (FC)             & 67.4 & 24.9k (1.0) & 24.0s (1.0) \\
Orchestration-v1          & 58.2 & 35.8k (1.4) & 62.4s (2.6) \\
Swarm-descriptive         & 70.4 & 47.3k (1.9) & 100.0s (4.2) \\
Swarm-none                & 65.4 & 52.8k (2.1) & 104.4s (4.4) \\
CSE-low                   & 59.8 & 58.5k (2.4) & 62.8s (2.6) \\
Swarm-full                & 70.5 & 67.2k (2.7) & 138.6s (5.8) \\
CSE-high                  & 73.1 & 74.0k (3.0) & 66.1s (2.8) \\
Orchestration-v2          & 75.6 & 117.3k (4.7) & 73.5s (3.1) \\
\bottomrule
\end{tabular}
\caption{Cost versus accuracy, ordered by tokens per task. Multipliers are
relative to the baseline. Orchestration-v2 attains the highest accuracy at the
highest token cost of any arm: giving every worker the full SOP and tool set plus a
re-dispatch round is what makes it both accurate and expensive.}
\label{tab:cost}
\end{table}

\begin{figure}[t]
\centering
\includegraphics[width=\columnwidth]{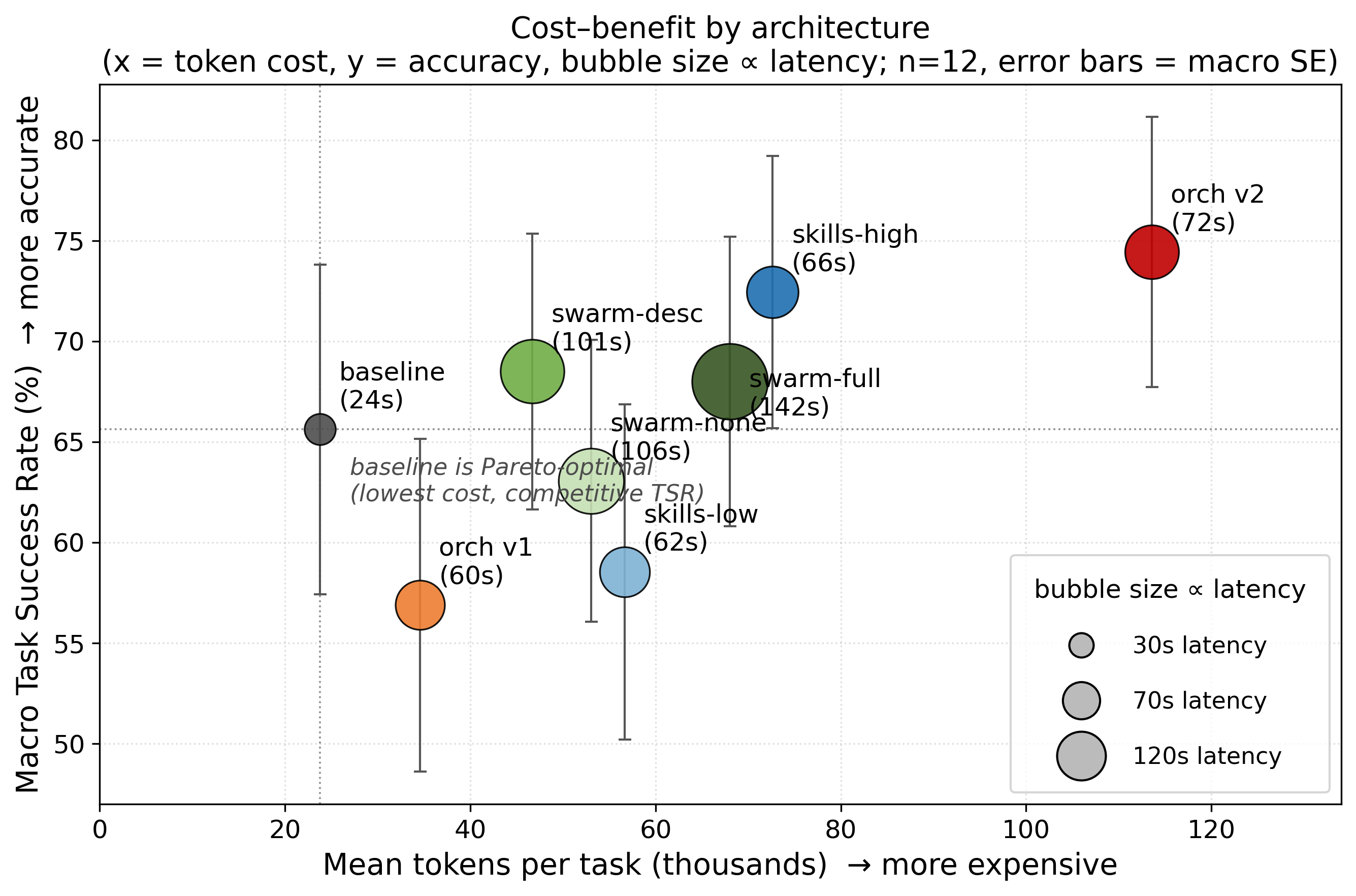}
\caption{Cost--benefit by architecture: mean token cost ($x$) versus macro TSR
($y$), bubble size proportional to latency; error bars are macro SE ($n{=}13$). No
arm sits above and to the left of the baseline; Orchestration-v2 reaches the top
accuracy at the highest token cost (\S\ref{sec:results-cost}).}
\label{fig:cost}
\end{figure}

The single-agent baseline is Pareto-optimal: cheapest on every axis yet competitive
on accuracy, with no arm both cheaper and more accurate (Figure~\ref{fig:cost}). Two
arms are strictly dominated (orchestration-v1, swarm-none), and the arms that edge
the baseline pay heavily for insignificant gains: CSE-high ($+5.7$ at $3.0\times$
tokens), swarm-descriptive ($+3.0$ at $1.9\times$), and swarm-full ($+3.1$ at
$2.7\times$). Orchestration-v2, the only arm to significantly beat the baseline
($+8.2$ points), is also the most token-expensive in the study ($4.7\times$ tokens,
$3.1\times$ latency): the changes that make it accurate (full SOP and tool set per
worker, a re-dispatch round) also make it costly, and at triple v1's tokens it does
not dominate v1 on cost. The reading is two-sided: the single agent is the efficient
default, and even the lone arm that separates from it buys that gain at the top of
the cost range. Appendix~\ref{sec:appendix-costofaccuracy} gives a per-SOP view.

\subsection{When does architecture matter?}
\label{sec:results-complexity}

Splitting the 13 SOPs at the median complexity $C_H$ (6 simple, 7 complex)
localizes where architecture matters (Table~\ref{tab:complexity}). On simple SOPs it
is a wash: the baseline is already strong ($71.8\%$). On complex SOPs the baseline
degrades most among capable arms ($-8.2$ points), while orchestration-v2 ($72.9\%$)
and CSE-high ($71.1\%$, drop-off only $-4.3$) hold up; each arm's edge over the
baseline tends to grow with complexity ($\rho$ up to $+0.38$), except the degraded
corpus, which collapses ($-22.5$, $\rho=-0.16$). No
correlation is significant ($n{=}13$ needs $|\rho|>0.56$), so we frame these as
workload-fit hypotheses: simple,
low-arity SOPs are architecture-insensitive, while long, branching SOPs are where
good skills and a corrected orchestrator most plausibly earn their cost.

\begin{table}[t]
\centering
\small
\begin{tabular}{lccc}
\toprule
Arm & Simple & Complex & $\rho_{\mathrm{edge}}$ \\
\midrule
Orchestration-v2  & 78.7 & 72.9 & $+0.31$ \\
CSE-high          & 75.4 & 71.1 & $+0.09$ \\
Swarm-full        & 72.6 & 68.8 & $+0.38$ \\
Swarm-descriptive & 72.8 & 68.4 & $+0.36$ \\
Swarm-none        & 67.4 & 63.7 & $+0.28$ \\
Baseline (FC)     & 71.8 & 63.6 & n/a \\
Orchestration-v1  & 60.4 & 56.4 & $+0.08$ \\
CSE-low           & 71.9 & 49.4 & $-0.16$ \\
\bottomrule
\end{tabular}
\caption{Macro TSR (\%) on the simple and complex halves (median split on human
complexity $C_H$, 6 SOPs each), and the Spearman correlation between each
arm's per-SOP edge over the baseline and $C_H$. Positive
$\rho_{\mathrm{edge}}$ means the arm's advantage grows with complexity; no $\rho$ is
significant at $n{=}13$ ($|\rho|>0.56$ needed for $p<0.05$).}
\label{tab:complexity}
\end{table}

\subsection{Ablation: open-tool skill guidance}
\label{sec:results-osg}

CSE scopes the executor to each skill's declared tools; Open-Tool Skill Guidance
(OSG) instead supplies the skills as advice but exposes \emph{all} tools
(\S\ref{sec:agent-configurations}). Ablating this on the two most tool-heavy SOPs
(Table~\ref{tab:osg}), tool scoping is itself workload-dependent: it helps when the
correct tools are few (KYB, 8 tools) but hurts on the 26-tool video-annotation SOP,
where open access recovers $+20$ points. Skill quality still dominates within OSG
($+28.9$ on KYB vs.\ $+4.8$ on video annotation).

\begin{table}[t]
\centering
\small
\resizebox{\columnwidth}{!}{%
\begin{tabular}{lccccc}
\toprule
SOP (\# tools) & FC & CSE-hi & CSE-lo & OSG-hi & OSG-lo \\
\midrule
KYB (8)            & 38.9 & \textbf{85.6} & 40.0 & 68.9 & 40.0 \\
Video annot.\ (26) & 34.4 & 37.6 & 33.6 & \textbf{57.6} & 52.8 \\
\bottomrule
\end{tabular}}
\caption{Open-tool (OSG) vs.\ scoped (CSE) skill execution on the two most
tool-heavy SOPs, TSR (\%), high/low skill corpora. Hard tool scoping (CSE) wins
when the relevant tools are few (KYB) but underperforms open access (OSG) on the
26-tool \texttt{video\_annotation}. Two SOPs; exploratory.}
\label{tab:osg}
\end{table}

\subsection{Evaluation reliability}
\label{sec:results-reliability}
 
Two scoring defects would flip conclusions if left uncorrected. A multi-field
output-contract bug truncated decisions into near-zero cells; fixing it for the
swarm lifted three SOPs (\texttt{aircraft\_inspection}, \texttt{email\_intent},
\texttt{patient\_intake}), and it is unverified for orchestration, whose multi-field
cells we therefore treat as provisional. A trace-denominator bug divided correct
answers by saved traces rather than tasks (inflating one cell to $97.9\%$ versus the
correct $93.5\%$); we compute every rate over the full task count. Both show that
outcome metrics need parse- and harness-level scrutiny before architectural
conclusions are drawn.
\section{Conclusion}
\label{sec:conclusion}

This study compares eight architectural configurations spanning four architecture
families across 12 shared
SOPs while holding the model, tasks, and tools fixed.
TSR effects vary sharply across SOPs, so aggregate means do not support a stable
architecture ranking. Architecture cost is clearer: every non-baseline arm uses
$1.5$--$4.8\times$ the tokens and $2.5$--$5.9\times$ the latency, making the
function-calling single agent the efficient default. Orchestration-v2 is the
one significant exception ($74.4\%$ versus $65.6\%$; odds ratio $1.78$, $p=0.008$),
yet also the most token-expensive arm; and its predecessor v1, an equally plausible
multi-agent implementation, lost to the single agent. Implementation quality can
thus outweigh the single- versus multi-agent choice: a poorly built multi-agent
system can underperform a good single agent.

There is no universal best architecture: the right choice is SOP-dependent, and added structure only plausibly pays off on complex, branching SOPs, a suggestive but not yet significant trend. Within the scope of enterprise SOPs, the practitioner message is direct: {\bf{start simple, measure marginal TSR on the target SOPs, repair
implementation failures before changing architecture families, and switch only
when the additional TSR justifies the additional architecture cost.
Architectural complexity should be earned by evidence, not assumed to be an
upgrade.}}

\section*{Limitations}
\label{sec:limitations}
Our scope is deliberately narrow. All results come from one benchmark, SOP-Bench,
whose enterprise themed SOPs and synthetic deterministic mock tools omit the
latency, errors, and nondeterminism of live services, so conclusions may not
transfer to production SOPs. Every arm runs on one model
(\texttt{claude-opus-4-8}); the baseline is stable across recent Opus versions
(Appendix~\ref{sec:appendix-baseline-models}), but we did not test other model
families, so we make no claim that the ranking, cost multipliers, or the
orchestration v1-to-v2 reversal generalize beyond it. With 12 SOPs the
between-SOP standard errors dominate, so most differences are not significant and
complexity correlations are directional only ($|\rho|>0.58$ needed at $n{=}12$); the
two-SOP OSG ablation is exploratory for the same reason. Each architecture is a
single implementation, sensitive to prompt and control-flow choices, and the
skill-quality contrast is a bundled degradation, so neither isolates the best
instantiation nor a single causal factor. We evaluate one run per task (binomial
uncertainty only, no stochastic-variance estimate), and although we recomputed
metrics from one task-count-correct source, arms span multiple runs and code
revisions. Finally, we report outcome and cost metrics only; the framework's
process metrics (procedural fidelity and LLM-judge reasoning/interaction quality)
need reference annotations absent from current traces and are left to future work.

\section*{Ethical Considerations}
This is a measurement study on a public benchmark and involves no new data
collection or human subjects. SOP-Bench is synthetic and industrially themed: the
account records, emails, and other fields are generated rather than drawn from real
users, so our experiments raise no direct privacy concerns.

The domains we study are nonetheless consequential. Several SOPs encode fraud and
referral-abuse enforcement, business verification, content moderation, and
healthcare-style intake, where an automated decision can suspend or close an
account, escalate a case, or gate access to a service. Our results show that task
success is well short of perfect and varies sharply by SOP, and that a poorly
specified skill can silently emit schema-violating or incorrect decisions while
appearing to run normally. We therefore caution against deploying LLM agents
autonomously on such procedures without human review, calibrated abstention, and
auditing. Our per-SOP, cost-aware evaluation is meant to support that deployment
judgment responsibly, not to endorse automating high-stakes enforcement.

Finally, coordinated architectures consume $1.5$ to $4.8$ times the tokens and $2.5$
to $5.9$ times the latency of a single agent, with a corresponding rise in compute
and energy use. Our finding that this overhead is usually not justified by accuracy
carries a positive efficiency implication: defaulting to the simplest sufficient
architecture lowers both operating cost and environmental footprint.

\newpage
\bibliography{custom}

\appendix
\section{Appendix: Skill-Definition Quality Rubric and Worked Examples}
\label{sec:appendix-rubric}

This appendix details the Skill-Definition Quality Rubric introduced in
\S\ref{sec:skill-quality-rubric} and illustrates it with a matched high- and
low-quality skill pair.

\subsection{Rubric}
\label{sec:appendix-rubric-def}

The rubric scores a skill definition on a 50-point scale across the five
dimensions in Table~\ref{tab:rubric-dims}. Scoring uses a two-stage protocol:
deterministic structural presence checks verify that the required fields exist
and are non-empty, and an LLM judge (\texttt{claude-sonnet-4}, temperature $0$,
one invocation per skill) rates the semantic quality of the content, so that a
definition is not credited merely for including an empty heading. Total scores
map to grades: Excellent ($\ge 90\%$), Good ($80$--$90\%$), Fair ($70$--$80\%$),
and Needs Improvement ($<70\%$).

\begin{table*}[t]
\centering
\small
\begin{tabular}{l c p{0.62\textwidth}}
\hline
\textbf{Dimension} & \textbf{Max} & \textbf{What it measures} \\
\hline
Identity \& discoverability & 10 & Name, description (what the skill does, when to use it, and the expected outcome), and usage scenarios that let a planner discover and select the skill. \\
Instruction quality & 15 & Ordered, specifically worded steps with per-step purpose and tool mapping, plus reasoning strategy, termination conditions, and constraints needed to execute the skill correctly. \\
Contract clarity & 10 & Preconditions, typed input and output schemas with examples, postconditions, and declared tool dependencies for reliable invocation and composition. \\
Disambiguation signals & 10 & Anti-patterns, explicit negative scope (what the skill does not do), and boundary constraints that prevent mis-selection against similar skills. \\
Governance metadata & 5 & Domain, complexity, risk level, and tags for cataloguing and lifecycle management. \\
\hline
\textbf{Total} & \textbf{50} & \\
\hline
\end{tabular}
\caption{The five dimensions of the Skill-Definition Quality Rubric and their maximum point allocations.}
\label{tab:rubric-dims}
\end{table*}

\subsection{Worked Examples}
\label{sec:appendix-rubric-ex}

We contrast the high- and low-quality versions of the same skill,
\texttt{assess-patient-risk} from the patient-intake SOP, which a planner uses to
compute a patient's lifestyle and overall risk levels. Both versions declare the
same two tools and differ only in how completely they document the procedure.
Table~\ref{tab:rubric-ex} gives the per-dimension scores.

The high-quality version scores 50/50 (Excellent). Its description states what
the skill does, when to use it, and what it returns, and explicitly bounds scope
(``it does NOT provide treatment recommendations''); it lists usage scenarios,
three ordered steps whose action and purpose are spelled out (lifestyle risk is
computed before overall risk), fully typed parameters with allowed values and
examples, an interpretation of each output field, and constraints that encode the
required ordering.

The low-quality version scores 33/50 (Needs Improvement). Degradation strips most
discovery and disambiguation signal while leaving the structured metadata intact,
so the loss concentrates in three dimensions. Identity and discoverability falls
from 10 to 6: the description collapses to a single generic sentence
(``Performs risk assessment on patient data to determine risk levels for clinical
purposes.'') with no when-to-use scenarios. Disambiguation signals falls from 10
to 2: the negative scope and boundary constraints are removed entirely.
Instruction quality falls from 15 to 11: the steps degrade to three vague bullets
(``Get lifestyle risk data. Process overall risk assessment. Compile results into
report.'') with no per-step action detail, tool mapping, or ordering. Contract
clarity (10 to 9) and governance metadata (5 to 5) are largely preserved, because
the typed input/output schema and catalogue metadata live in the structured
definition that degradation leaves untouched. The net effect is a 17-point drop
(from $100\%$ to $66\%$), concentrated in the procedural and disambiguation
content, and it is this kind of separation that the corpus-level manipulation
check in \S\ref{sec:skill-quality-rubric} summarizes.

\begin{table}[t]
\centering
\small
\begin{tabular}{l c c}
\hline
\textbf{Dimension} & \textbf{High} & \textbf{Low} \\
\hline
Identity \& discoverability & 10 & 6 \\
Instruction quality & 15 & 11 \\
Contract clarity & 10 & 9 \\
Disambiguation signals & 10 & 2 \\
Governance metadata & 5 & 5 \\
\hline
Total & 50 & 33 \\
Percentage & 100\% & 66\% \\
Grade & Excellent & Needs Impr. \\
\hline
\end{tabular}
\caption{Per-dimension rubric scores for the high- and low-quality versions of \texttt{assess-patient-risk}.}
\label{tab:rubric-ex}
\end{table}

\section{Appendix: Per-SOP Skill-Quality Results}
\label{sec:appendix-skillquality}

Table~\ref{tab:skill-per-sop} gives the per-SOP task success rates for the
high- and low-quality skill corpora (CSE-high, CSE-low), with the function-calling
baseline for reference, underlying the aggregate skill-quality comparison in
\S\ref{sec:results-knobs}. The high$-$low gap is concentrated in a few long,
branching procedures; on most SOPs the two corpora are within a few points, and on
several the degraded corpus is marginally higher.

\begin{table}[t]
\centering
\small
\begin{tabular}{l c c c}
\hline
\textbf{SOP} & \textbf{Base.} & \textbf{CSE-hi} & \textbf{CSE-lo} \\
\hline
aircraft\_inspection   & 22.3 & 43.8 & 49.1 \\
content\_flagging      & 28.6 & 36.9 & 37.5 \\
customer\_service      & 88.5 & 80.8 & 74.4 \\
dangerous\_goods       & 91.6 & 77.0 & 89.8 \\
email\_intent          & 86.6 & 90.9 & 51.6 \\
know\_your\_business   & 38.9 & 85.6 & 40.0 \\
patient\_intake        & 93.9 & 93.9 & 93.9 \\
referral\_abuse\_v1    & 84.5 & 97.0 & 95.5 \\
referral\_abuse\_v2    & 100.0 & 93.5 &  2.5 \\
traffic\_spoofing      & 76.0 & 78.0 & 78.0 \\
video\_annotation      & 34.4 & 37.6 & 33.6 \\
video\_classification  & 78.7 & 85.3 & 79.7 \\
warehouse\_package     & 52.0 & 50.0 & 51.3 \\
\hline
\textbf{Macro} & \textbf{67.4} & \textbf{73.1} & \textbf{59.8} \\
\hline
\end{tabular}
\caption{Per-SOP TSR (\%) for the function-calling baseline and the high- and
low-quality skill corpora (CSE-high, CSE-low). The macro row is the mean over the
13 SOPs (equal weight per SOP).}
\label{tab:skill-per-sop}
\end{table}

\section{Appendix: Per-SOP Cost of Accuracy}
\label{sec:appendix-costofaccuracy}

\begin{figure*}[t]
\centering
\includegraphics[width=0.85\textwidth]{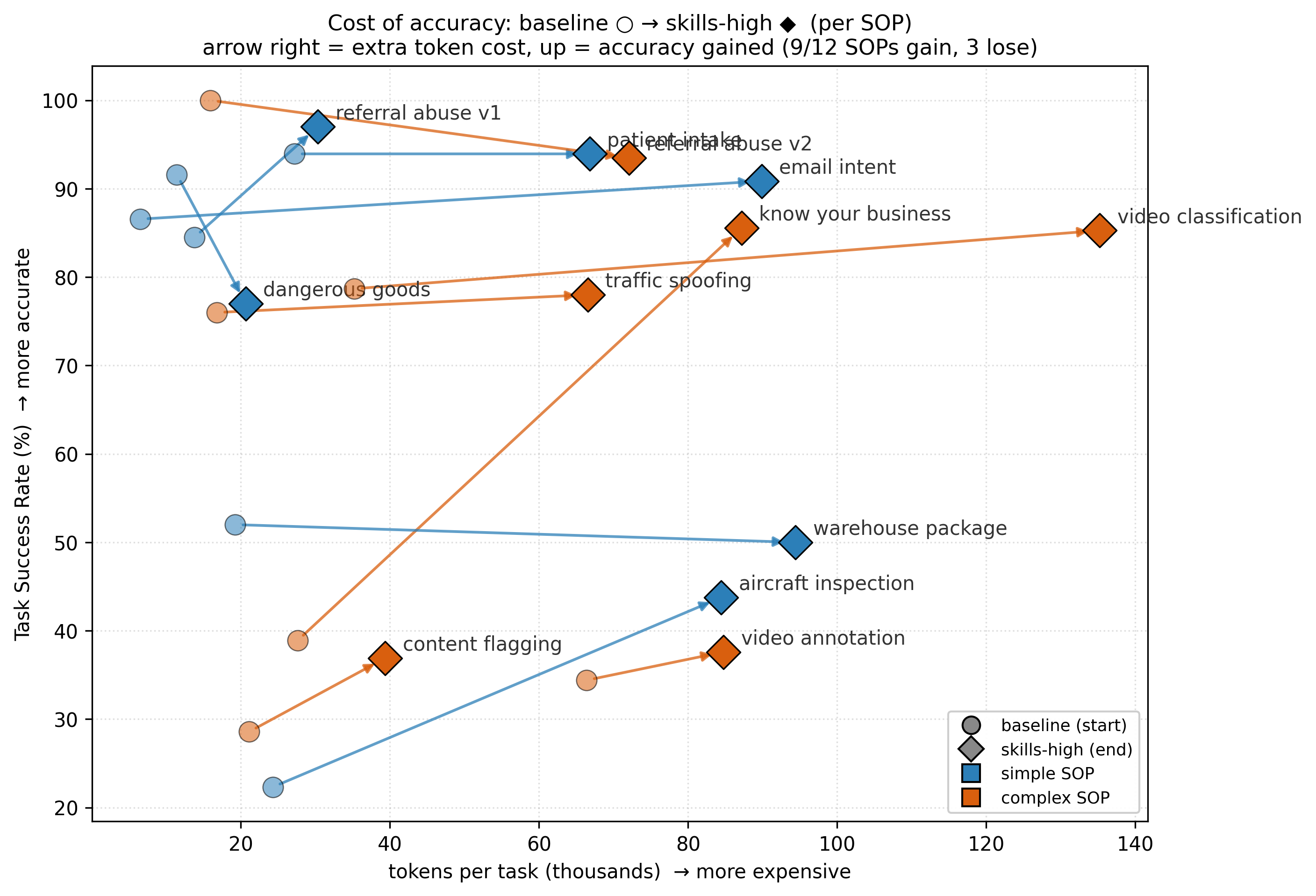}
\caption{Per-SOP cost of accuracy, baseline ($\circ$) to CSE-high ($\diamond$),
colored by SOP complexity. Rightward displacement is extra token cost; upward is
accuracy gained.}
\label{fig:costarrows}
\end{figure*}

Figure~\ref{fig:costarrows} traces, for each SOP, the move from the single-agent
baseline to the highest-cost skill configuration (CSE-high): rightward is extra
token cost, upward is accuracy gained. It makes concrete the aggregate cost story
of \S\ref{sec:results-cost}: the arrows point sharply right on nearly every SOP
(a large, consistent token cost) while moving little vertically, and on a few
SOPs they point down (extra cost, lower accuracy). We use CSE-high as an
illustrative high-cost skill configuration with complete per-SOP cost data.

\section{Appendix: Random-Effects Model for the Baseline Comparison}
\label{sec:appendix-randomeffects}

This appendix specifies the confirmatory single-arm-versus-baseline test summarized
in \S\ref{sec:evaluation}. We treat each SOP as one study and combine the 13 per-SOP
effects with a random-effects meta-analysis \citep{dersimonian1986meta}, which is the
natural model when the quantity of interest is an arm's effect on a \emph{new} SOP
and the dominant uncertainty is between-SOP rather than between-task
(\S\ref{sec:results-sop}).

\paragraph{Per-SOP effect size.}
For SOP $i$ let the arm answer $a_i$ of $n_i$ tasks correctly and the baseline
$b_i$ of $n_i$. We use the log odds ratio
\[
y_i \;=\; \log\!\frac{a_i\,(n_i-b_i)}{b_i\,(n_i-a_i)},
\]
\[
v_i \;=\; \frac{1}{a_i}+\frac{1}{n_i-a_i}+\frac{1}{b_i}+\frac{1}{n_i-b_i},
\]
with $v_i$ the standard large-sample variance of $y_i$; a $0.5$ continuity
correction is applied to any zero cell. A positive $y_i$ means the arm beats the
baseline on SOP $i$.

\paragraph{Random-effects pooling.}
We model $y_i = \theta + u_i + \varepsilon_i$ with $\varepsilon_i \sim
\mathcal{N}(0, v_i)$ the within-SOP sampling error and $u_i \sim \mathcal{N}(0,
\tau^2)$ the between-SOP heterogeneity, so $\theta$ is the mean log odds ratio
across SOPs. The heterogeneity variance $\tau^2$ is estimated by the
method-of-moments DerSimonian--Laird estimator \citep{dersimonian1986meta} from
Cochran's $Q$, and the pooled estimate is the inverse-variance weighted mean
\[
\hat\theta \;=\; \frac{\sum_i w_i\, y_i}{\sum_i w_i},
\qquad
w_i \;=\; \frac{1}{v_i + \hat\tau^2}.
\]
Weighting by $v_i + \hat\tau^2$ (rather than $v_i$ alone) is what prevents a single
high-task SOP from dominating: once between-SOP variance is nonzero, additional
tasks on one SOP cannot shrink its weight below $\approx 1/\hat\tau^2$.

\paragraph{Small-sample inference (Knapp--Hartung).}
With only $k=13$ SOPs, the usual Wald test that treats $\hat\theta/\sqrt{1/\sum_i
w_i}$ as standard normal is anticonservative. We therefore use the
Knapp--Hartung adjustment \citep{knapp2003improved}, which rescales the variance by
the weighted residual mean square
\[
q \;=\; \frac{1}{k-1}\sum_i w_i\,(y_i-\hat\theta)^2,
\]
and refers the statistic $\hat\theta \,/\, \sqrt{q \, / \sum_i w_i}$ to a
$t$-distribution with $k-1$ degrees of freedom. For few studies this method
controls the type-I error rate far better than the DerSimonian--Laird--Wald
combination \citep{inthout2014hartung}, at the cost of wider, more honest intervals.
We report the pooled odds ratio $\exp(\hat\theta)$, its $t$-based confidence
interval, and the corresponding $p$-value; this is the test behind the single
significant separation in \S\ref{sec:results-orch} (orchestration-v2 vs.\ baseline,
odds ratio $1.72$, $p=0.007$).

\section{Appendix: Function-Calling Baseline across Claude Opus Versions}
\label{sec:appendix-baseline-models}

We evaluate the function-calling baseline with Claude Opus 4.6, 4.7, and 4.8 on
the same 13 SOPs. Each model has one valid v1 trace for every task
($2{,}124$ traces per model) and $100\%$ execution completion. Table~\ref{tab:baseline-models}
reports per-SOP task success rate (TSR) and task-weighted resource means computed
directly from those traces. The equal-SOP macro TSRs (mean $\pm$ SE) are $68.5\pm9.3\%$,
$66.5\pm8.0\%$, and $67.4\pm7.7\%$ for Opus 4.6, 4.7, and 4.8,
respectively; no model is uniformly best across SOPs.

\begin{table*}[t]
\centering
\small
\setlength{\tabcolsep}{3.5pt}
\resizebox{\textwidth}{!}{%
\begin{tabular}{l rrr rrr rrr}
\toprule
& \multicolumn{3}{c}{\textbf{Claude Opus 4.6}}
& \multicolumn{3}{c}{\textbf{Claude Opus 4.7}}
& \multicolumn{3}{c}{\textbf{Claude Opus 4.8}} \\
\cmidrule(lr){2-4}\cmidrule(lr){5-7}\cmidrule(lr){8-10}
\textbf{SOP} & \textbf{TSR} & \textbf{Tok.} & \textbf{Lat.}
& \textbf{TSR} & \textbf{Tok.} & \textbf{Lat.}
& \textbf{TSR} & \textbf{Tok.} & \textbf{Lat.} \\
& (\%) & (k) & (s) & (\%) & (k) & (s) & (\%) & (k) & (s) \\
\midrule
Aircraft inspection   & 0.9   & 22.5  & 29.1 & 8.9   & 23.3 & 20.0 & 22.3  & 24.3 & 31.1 \\
Content flagging      & 33.3  & 19.6  & 31.9 & 36.3  & 21.6 & 16.8 & 28.6  & 21.1 & 20.8 \\
Customer service      & 89.1  & 33.1  & 21.0 & 92.3  & 42.6 & 14.1 & 88.5  & 38.0 & 23.0 \\
Dangerous goods       & 93.8  & 6.3   & 12.9 & 90.5  & 12.9 & 11.3 & 91.6  & 11.4 & 15.0 \\
Email intent          & 98.9  & 5.9   & 10.7 & 72.0  & 6.9  & 5.9  & 86.6  & 6.5  & 9.8  \\
Know your business    & 60.0  & 23.2  & 28.5 & 48.9  & 28.4 & 20.6 & 38.9  & 27.6 & 33.0 \\
Patient intake        & 90.9  & 23.6  & 33.1 & 95.5  & 32.7 & 21.7 & 93.9  & 27.2 & 29.3 \\
Referral abuse v1     & 95.5  & 10.6  & 15.0 & 81.5  & 12.7 & 11.2 & 84.5  & 13.8 & 18.6 \\
Referral abuse v2     & 100.0 & 17.1  & 19.7 & 100.0 & 15.0 & 12.3 & 100.0 & 15.9 & 20.6 \\
Traffic spoofing      & 79.0  & 13.1  & 21.5 & 69.0  & 10.1 & 10.3 & 76.0  & 16.8 & 19.9 \\
Video annotation      & 16.0  & 100.8 & 34.4 & 28.8  & 42.1 & 13.9 & 34.4  & 66.4 & 37.2 \\
Video classification  & 81.2  & 31.6  & 31.4 & 88.3  & 34.1 & 18.7 & 78.7  & 35.2 & 35.0 \\
Warehouse inspection  & 51.3  & 16.1  & 22.9 & 52.7  & 20.2 & 12.4 & 52.0  & 19.2 & 18.2 \\
\midrule
\textbf{Mean TSR $\pm$ SE}
& $\mathbf{68.5 \pm 9.3}$ & -- & --
& $\mathbf{66.5 \pm 8.0}$ & -- & --
& $\mathbf{67.4 \pm 7.7}$ & -- & -- \\
\bottomrule
\end{tabular}}
\caption{Per-SOP function-calling baseline metrics for three Claude Opus
versions. Tok. is mean total tokens per task in thousands; Lat. is mean wall-clock
latency per task in seconds. The final row reports the equal-SOP TSR mean $\pm$
standard error; token and latency cells are shown as dashes. All values are
computed from the common trace set. Each task has one saved outcome at temperature
1, so the table is a model-version sensitivity analysis rather than a stochastic
ranking. Latency is descriptive because runs occurred at different times and no
service-load metadata was logged.}
\label{tab:baseline-models}
\end{table*}

The comparison also has format-sensitive multi-field SOPs and does not record an
immutable parser revision in each run. We therefore use the table to show that
the baseline conclusion is not unique to Opus 4.8, not to claim a general model
ordering. The complete trace-derived values, including correctness counts, ECR,
LLM calls, tool calls, and turns, are generated by
\path{analysis/baseline_model_comparison.py} and stored in
\path{analysis/baseline_model_comparison.csv}.

\section{Appendix: Agent Configurations}
\label{sec:appendix-configs}

This appendix gives the full control flow of the eight arms summarized in
Table~\ref{tab:arms}.

\paragraph{Function-calling baseline.}
The baseline is a single agent that receives the full SOP, task input, and
benchmark tool schemas. It alternates between model-generated structured tool
calls and deterministic tool responses, then emits a tagged final decision. The
baseline provides a single-agent reference with unrestricted access to the
benchmark's declared tools.

\paragraph{Constrained Skill Execution (CSE).}
CSE represents skills as reusable procedural artifacts. For each task, a selector
first receives the SOP, task, and a catalog of skill identities, descriptions,
and declared tool dependencies. It chooses one or more relevant skills. The
executor then processes selected skills in model-chosen order, supplying the
skill's Markdown instructions and exposing only the tools declared for that
skill. CSE tests skills as both procedural guidance and explicit
capability boundaries.

\paragraph{Open-Tool Skill Guidance (OSG).}
OSG is a distinct formulation, not a variant to pool with CSE. It uses the same
high- and low-quality skill corpora and gives the executor each skill's
instructions, recommended tools, contraindications, anti-patterns, and exit
criteria, but unlike CSE it exposes all benchmark tools: calls outside a skill's
recommended set execute and are logged as deviations rather than rejected. We
evaluate OSG on only two complex SOPs, as a focused study of skills as advisory
guidance under broader agent autonomy rather than a benchmark-wide comparison.

\paragraph{Centralized multi-agent orchestration.}
A decomposer, worker, synthesizer pipeline: a planning call partitions the SOP into
subtasks with per-subtask tool subsets and declared dependencies; each worker sees
only its assigned SOP section, task context, and tool subset, plus prior workers'
summaries for dependencies; and a synthesizer combines the worker summaries and tool
outputs into the final decision. The evaluated implementation runs dependency-ready
workers serially rather than in parallel. We call this orchestration-v1.

\paragraph{Orchestration-v2.}
A trace-level diagnosis traced most of v1's losses to one design gap: because each
worker sees only its slice and no component may resolve a missing input, a blocked
worker can only abstain, so the pipeline emits placeholder decisions.
Orchestration-v2 rectifies this with five changes: every worker receives the full
SOP and tool set; a commit-with-basis contract forbids placeholder outputs; the
synthesizer may resolve residual gaps; a single re-dispatch round lets a blocked
worker request what it is missing; and each decision is validated against the SOP's
permitted values. We treat v1 and v2 as distinct arms.

\paragraph{Triage-guided swarm.}
A triage agent partitions the SOP into specialists with exclusive tool sets. Each
specialist receives its SOP phase, permitted tools, a shared-blackboard summary of
prior findings, and peer descriptions, then returns findings and may hand off to
another specialist. Revisits are allowed, execution is capped at ten handoffs, and
missing or invalid handoffs route to a decision agent, which sees the full SOP and
blackboard before producing the final decision. This is thus a triage-guided
specialist-handoff design, not an unconstrained peer-to-peer one. The swarm has
three \texttt{blueprint\_level} settings: \texttt{none} (no shared blueprint);
\texttt{descriptive} (a common goal and unordered high-level stages, with no
tool-level instructions, thresholds, branch logic, or enforced ordering); and
\texttt{full} (a recommended stage order and routing logic). Because \texttt{full}
combines centralized planning with decentralized execution, we treat it as a hybrid
condition rather than pooling it with the pure swarm settings.

\section{Appendix: Why CSE-low Collapses on referral\_v2}
\label{sec:appendix-cselow-v2}

CSE-low scores near zero on \texttt{referral\_abuse\_v2} ($2.5\%$), far below its
own performance elsewhere. Trace inspection shows this is a skill-quality failure
that surfaces as an output-schema violation, not a metrics bug. The SOP requires the
final answer in a field named \texttt{final\_decision}. The degraded skill drops
that output contract, so the agent computes its decision but files it under a
concocted key: in $194$ of $200$ traces the decision is placed under
\texttt{enforcement\_action} (a name the low-quality skill invented), whereas
CSE-high reports it under the SOP-mandated \texttt{final\_decision} in $179$ of
$187$ traces. Because the scorer reads \texttt{final\_decision}, every mis-keyed
answer is counted wrong.

The underlying decisions are largely correct: read from the misplaced field, the
predicted action matches ground truth on $193$ of $200$ tasks ($\approx 96\%$),
comparable to CSE-high. This is distinct from the baseline
\texttt{referral\_v2} case (\S\ref{sec:results-reliability}), which was a
metrics-side parsing bug we corrected: there the agent emitted the answer in the
required field and only the offline extractor mis-read it. Here the agent itself
violates the output contract because the skill failed to state it, so the failure is
a genuine consequence of skill quality and we report it as-is. It is a concrete
example of how a poor skill degrades an agent silently: the reasoning is sound but
the output is unusable to any downstream consumer that expects the specified schema.

\end{document}